\documentclass[11pt,a4paper]{article}
\usepackage[hyperref]{acl2019}
\usepackage{times}
\usepackage{latexsym}
\usepackage{multirow}
\usepackage{graphicx}

\usepackage{url}

\aclfinalcopy 

\title{Data Ordering Patterns for Neural Machine Translation: \\ An Empirical Study}

\author{Siddhant Garg \\
  University of Wisconsin-Madison \\
  \texttt{sgarg33@wisc.edu}}
\date{}

\begin{document}
\maketitle
\begin{abstract}
Recent works show that ordering of the training data affects the model performance for Neural Machine Translation. Several approaches involving dynamic data ordering and data sharding based on curriculum learning have been analyzed for the their performance gains and faster convergence. In this work we propose to empirically study several ordering approaches for the training data based on different metrics and evaluate their impact on the model performance. Results from our study show that pre-fixing the ordering of the training data based on perplexity scores from a pre-trained model performs the best and outperforms the default approach of randomly shuffling the training data every epoch. 
\end{abstract}

\section{Introduction}
Sequence to sequence  (seq2seq) learning models \cite{NIPS2014_5346} using an encoder decoder framework have been a popular choice for the task of machine translation before the recent popularity of transformer models \cite{NIPS2017_7181}. Several recent works \cite{bahdanau2014neural,DBLP:journals/corr/LuongPM15,Zhao2018AttentionviaAttentionNM,shankar-etal-2018-surprisingly} aims to improve the translation performance of seq2seq models by formulating new attention mechanisms to capture similarity between the encoder and decoder states. Recent works \cite{zhang-etal-2017-boosting}\cite{van-der-wees-etal-2017-dynamic} \cite{wang-etal-2018-dynamic}  suggests that apart from attention mechanisms, data ordering patterns also affect the performance of neural machine translation. In this work we aim to empirically analyze the performance improvements of different data ordering patterns of the training data on the English-Vietnamese translation task.

Curriculum learning \cite{Bengio:2009:CL:1553374.1553380} proposes that choosing an ordering of the training samples from easier to learn examples to harder to learn examples can help train better models and achieve faster convergence. We take insights from curriculum learning and evaluate approaches to rank the training data based on complexity using several different metrics.

We empirically observe that a pre-fixed data ordering pattern based on sorted perplexity scores from a pre-trained model is able to outperform the default approach of randomly shuffling data every epoch and gets a 1.7 BLEU score improvement. We also perform experiments to analyze the effect of these ranking metrics from different pre-trained models and conclude that a shallow architecture suffices to achieve performance gains by providing an efficient data ordering pattern.

The remaining paper is structured by first describing the related work, followed by the different proposed data ordering patterns and experimental results and concluding by briefly describing future work extensions and conclusion from this study.

\section{Related Work}
Curriculum learning has been studied and applied to various Machine Learning and Natural Language Processing tasks \cite{tsvetkov-etal-2016-learning} \cite{DBLP:journals/corr/CirikHM16} \cite{pmlr-v70-graves17a}. Most training protocols for Neural Machine Translation randomize the order of sentence pairs in the training corpus \cite{sennrich-etal-2017-nematus} \cite{DBLP:journals/corr/abs-1712-05690}. One of the initial studies carried out by \newcite{kocmi-bojar-2017-curriculum} proposed a curriculum learning based mini bucketing approach using sequence length, number of coordinating conjunctions and word ranks by ensuring that samples within each mini-batch have similar linguistic properties. They show that translation quality can be improved by presenting samples from easy to hard based on sentence length and vocabulary frequency.

\newcite{zhang-etal-2017-boosting} propose a data boosting and bootstrap paradigm using a probabilistic approach which assigns higher weights to training examples that have lower perplexities in previous epoch. Similarly, \newcite{van-der-wees-etal-2017-dynamic,wang-etal-2018-dynamic} improve the training efficiency of NMT by dynamically selecting different subsets of training data between different epochs using domain relevance and difference between the training costs of two iterations respectively.

\newcite{DBLP:journals/corr/abs-1811-00739} propose to split the training samples into a predefined number of bins based on varied difficulty metrics like maximum and average word frequency rank. \newcite{DBLP:journals/corr/abs-1903-09848} uses a difficulty and competence based metric for faster convergence and better performance than uniformly sampling training examples.

Our approach differs from these recent works in the main aspect that the model can access the entire training data in each epoch in our approach as compared to other techniques which partition the training data set and provide a different portion of the training data to the model each epoch. 

\section{Data Ordering Patterns}
Neural Machine Translation using seq2seq learning models uses mini-batches of data for training and typically requires multiple passes over the training data to reach convergence. The default baseline strategy used for training the models is using a random shuffle of the training data every epoch. 

Traditional curriculum learning approaches propose breaking the training data into groups based on the training data complexity and using these groups of training order in a sequential order from lower to higher complexity. Our approach just re-orders the training data and the model can access the whole training data in every epoch as contrasted to curriculum learning where it accesses specific portions of the training data in different training epochs. 

Each train data point is a pair of sentences: one in the source and the other is the corresponding translation in the target language. We propose 4 different ordering strategies for the training data where this order is fixed before the training starts and mini-batches are chosen from this ordered training data sequentially. No random shuffling of training data is carried out every epoch when using these data patterns. We do not disturb the sentences within a pair in any of our strategies, but rather just re-order different pairs. The following data patterns are proposed:

\begin{itemize}
    \item \textbf{Random Shuffle:} Randomly shuffle the training data points. This results in a model which shuffles the training data only once before training starts.
    \item \textbf{Sequence Length Order:} Sort the training data based on the length of the sentences of the source or target language. There are 2 orders hence obtained: sorted in ascending and descending order of lengths for each of the source and target sentences.
    \item \textbf{Perplexity based Order:} Sort the training data based on perplexity scores of each training data pairs from a pre-trained model $M$. For the purpose of optimization, if the cross-entropy of sentence pairs is denoted by $H(p)$, the perplexity is defined as $2^{H(p)}$. This results in 2 orderings: sorted in ascending and descending order of perplexity scores.
    \item \textbf{BLEU based Order:} Sort the training data based on BLEU score \cite{Papineni:2002:BMA:1073083.1073135} of each training data pairs from a pre-trained model $M$. This results in 2 orderings: sorted in ascending and descending order of BLEU scores
\end{itemize}

For the perplexity and BLEU score based ordering, we use a pre-trained model on the same training corpus. We experiment with using different pre-trained models to analyze the impact on the performance. The experiments and results are presented in the next section. 

\section{Experiments}

\begin{table*}[h]
\begin{centering}
\begin{tabular}{|c|c|c|c|}
\hline
Data Ordering Pattern            & Epochs & Test PPL & Test BLEU \\ \hline
Random Shuffle every epoch       & 32       & 16.55          & \textbf{18.1} \\ \hline
Random Shuffle once                 & 30        & 18.78        & 18.0 \\ \hline
Ascending Sequence Length Order for source language  & 15             &   24.02       & 16.6 \\ \hline
Descending Sequence Length Order for source language & 31             &    20.31       & 17.1 \\ \hline
Ascending Sequence Length Order for target language & 23            &   29.64         & 15.6 \\ \hline
Descending Sequence Length Order for target language & 29          &    23.66          & 16.8 \\ \hline
Ascending PPL Order (From Pre-trained base model)       & 32       &   14.69              & \textbf{19.8} \\ \hline
Descending PPL Order (From Pre-trained base model)      & 31        &   15.28             & 19.1 \\ \hline
Ascending BLEU Order (From Pre-trained base model)            & 31         &    15.46           & 18.9 \\ \hline
Descending BLEU Order (From Pre-trained base model)          & 29           &   15.78          & 18.6 \\ \hline
\end{tabular}
\caption{Effect of different data ordering patterns on the performance on the IWSLT Vietnamese test set. Random shuffle every epoch refers to the default approach of randomly shuffling the data every epoch. Random Shuffle once refers to randomly shuffling the data once before the training starts. PPL refers to perplexity. Epochs refers to number of epochs required for convergence. }
\label{table:results}
\end{centering}
\end{table*}

\subsection{Training Details and Hyper-parameters} \label{models}
We use an encoder-decoder architecture with Bahdanau attention \cite{bahdanau2014neural} using two layers of 512 units of LSTM encoder and decoder with a 0.2 dropout. We refer to this as the base model when presenting the results. We also experiment with a smaller architecture of the encoder-decoder framework using just 1 layer of 128 unit LSTM encoder and two layers of 128 unit LSTM decoder without attention having a 0.2 dropout probability and refer to this as the small model. The model was trained using an Adam \cite{Kingma2014AdamAM} optimizer with a learning rate of ${10}^{-5}$ . Training was performed on a 12GB Titan-X GPU using a batch size of 128. We use the BLEU score on the test data as the metric to evaluate the performance. 

\subsection{Dataset} \label{dataset}
For our experiments we used the standard IWSLT 2015 English-Vietnamese language data set \cite{Cettolo2015TheI2} which has
around 133k training sentence pairs. We sample 60k training samples from the training data by filtering out duplicates and sentences of sequence lengths more than 60 and less than 5 to have a consistent evaluation between the different data shuffling patterns. We use the original validation and test data splits for the experiments.

Since our experiments aim to empirically contrast the performance of the different data ordering strategies, we do not use an ensemble of seq2seq models and use greedy decoding instead of beam-decoding (which adds to the memory consumed and the training time of the model). Hence our BLEU scores are typically lower than the state of the art performance of 26.1 on this task.

\subsection{Results}
We present the experimental results using different data ordering patterns of the training data in table \ref{table:results}. Randomly shuffling the data once before training can achieve a comparable BLEU score to the default technique used for seq2seq model training of randomly shuffling the training data every epoch, but the model with shuffling every epoch achieves a considerably lower test perplexity than with a single shuffle of the training data. A simple curriculum learning based approach of sorting the source language sentences based on their lengths performs considerably inferior to a randomly sorted ordering. Sorting based on the target language sentences length performs better than sorting based on the source language sentences lengths but is still not comparable to the random shuffling performance. This can possibly because of the optimizer getting stuck at a local optima and converging earlier rather than finding the global optima as is evidenced by the small number of epochs required for convergence.

Using data ordering patterns sorted on metrics like perplexity and BLEU outperforms the default approach of randomly shuffling the data. Perplexity and BLEU are the most commonly used estimators of the complexity of a sentence pair to be translated correctly by a translation model. A sentence pair having a lower perplexity or BLEU score than another from a pre-trained model implies that the first sentence pair was easier for the model to translate than the latter. From the table it is empirically observed that using an ascending order sorted approach performs the best and gets an improvement in BLEU score of 1.7 points from the default setting. We conjecture this is because the model first accesses less complex examples followed by more complex examples in line with the idea of curriculum learning.

The interesting observation made here is that a descending order sorted training data schedule based on perplexity or BLEU also outperforms the default setting of random shuffling though the model accesses training data samples from more complex to less complex. This idea is slightly in contrast to the curriculum learning approach of providing the model with training examples in increasing order of complexity.

\subsection{Comparison across models} \label{compare_models}

We evaluate the performance of the data ordering sorted based on perplexity and BLEU score from 2 different pre-trained models and show empirically that a smaller trained model is able to provide comparable performance gains to a larger trained model for these 2 data ordering strategies. The base and small models have been explained in section \ref{models}. The results are presented in table \ref{table:diff_models}

From these results, we can infer that even a smaller trained model can give a good estimate of the training data sentence pair complexities through perplexity and BLEU metrics. Also this shows that the improvements in performance due to these specific data orderings can be generalized across different models. We also note that perplexity based ascending order is the best performing approach. 

\begin{table}[h]
\resizebox{\linewidth}{!}{
\begin{tabular}{|c|c|c|c|}
\hline
Pre-Trained Model            & Data Pattern       & Epochs & BLEU \\ \hline
\multirow{4}{*}{Small Model} & Asc PPL   & 31            & 19.7 \\ \cline{2-4} 
                             & Desc PPL  & 30            & 19.1 \\ \cline{2-4} 
                             & Asc BLEU         & 32            & 18.9 \\ \cline{2-4} 
                             & Desc BLEU        & 31            & 18.5 \\ \hline
\multirow{4}{*}{Base Model}  & Asc PPL   & 32            & 19.8 \\ \cline{2-4} 
                             & Desc PPL  & 31            & 19.1 \\ \cline{2-4} 
                             & Asc BLEU         & 31            & 18.9 \\ \cline{2-4} 
                             & Desc BLEU        & 29            & 18.6 \\ \hline
\end{tabular}
}
\caption{Comparing performance of data ordering patterns based on sorted perplexity and BLEU score from 2 different pre-trained models. Here PPL, Asc and Desc refer to perplexity, ascending order and descending order respectively. Epochs refers to number of epochs required for convergence.}
\label{table:diff_models}
\end{table}

\section{Ongoing and Future Work}
Ongoing work includes verifying this conjecture on the seq2seq framework for other machine translation datasets like English-German and English-French datasets. An extension of this work is to empirically verify if using a transformer based architecture can still provide similar gains from using a perplexity sorted ordering of the training data from a pre-trained model.

Results from section \ref{compare_models} show that a small model can be used to rank the training data points and this can then be used for improving the performance of a model trained on this data. This shows a promising future direction of work of making a NMT pipeline involving  a low-resource model which can be trained fast and in a computationally inexpensive way for ranking the training data points and a larger model which can exploit this ranked order and produce performance improvements for machine translation. 

\section{Conclusion}
From this study, we conclude that data ordering patterns can have an effect on the model performance for neural machine translation. While simple heuristics like sentence length actually lead to a drop in model performance, heuristics specific to evaluating NMT performance like perplexity and BLEU score can be good measures to rank the data for training and get performance gains over the default approach of randomly sampling data points for training.

\bibliography{acl2019}
\bibliographystyle{acl_natbib}

\end{document}